# Enhancing the Forecasting Capability of Multi-Model Blending Algorithms for Extreme Precipitation via Joint Use of Station and Gridded Observations

Authors: Yu Wang (王玉)[a], Yong Cao (曹勇)[a,b†], Kan Dai (代刊)[a], Yue Shen (沈悦)[c], Xiaoqing Zeng (曾晓青)[a], Ruixia Zhao (赵瑞霞)[a],

[a]National Meteorological Center of China Meteorological Administration, Beijing, 100081, China.

[b]Xiong'an Institute of Meteorological Artificial Intelligence, Xiong'an New Area, 070001, China.

[c]China Meteorological Administration Training Center, Beijing, 100081, China.

Corresponding author: Yong Cao (caoyong@cma.gov.cn)

**Abstract**

Accurate extreme precipitation forecasting is critical for disaster mitigation but remains challenging for numerical weather prediction (NWP) models due to systemic intensity underestimation and spatial displacement. Traditional precipitation multi-model blending algorithms perform pixel-by-pixel blending on the forecast field based on weights, which may lead to the expansion of precipitation areas and the smoothing of extreme values. This study proposes an U-Net based two-stage framework: probability classification followed by value reconstruction, to blend forecasts from six major NWP models. A novel station-grid joint supervision mechanism is introduced by integrating observations from 2411 national meteorological stations in China into the loss function, simultaneously constraining spatial structures and peak intensities.

Evaluations using independent samples from the 2025 flood season demonstrate that our model significantly outperforms both individual NWPs and current operational products. For rainstorms (≥50 mm), the Threat Score (TS) improved by 38.4% compared to the best NWP. Notably, for extreme events (≥100 mm) driven by extratropical cyclones and the subtropical high, the model successfully elevated the TS to above 0.1, transforming forecasts from having negligible reference value into those with certain operational utility. Furthermore, the model exhibits data-driven spatial correction capabilities, effectively realigning systematic rainbelt displacements with actual precipitation centers. The inclusion of station observations specifically enhanced the TS for rainstorms by 10.4% and effectively balanced the Bias. These results highlight the efficacy of multi-source joint supervision in enhancing the capture of extreme precipitation events.

## 1. Introduction

With global climate warming, the frequency and intensity of extreme weather events have increased significantly. Secondary disasters triggered by extreme precipitation, such as floods and landslides, pose a serious threat to socio-economic development and the safety of people's lives and property. Therefore, improving the accuracy of extreme precipitation forecasting is not only an urgent need for the advancement of meteorological science, but also a key component of disaster prevention and mitigation efforts (Busker, van den Hurk et al. 2025).

Current precipitation forecasting primarily relies on Numerical Weather Prediction (NWP). Although, global models represented by ECMWF (European Center for Medium-Range Weather Forecasts) and NCEP-GFS (National Centers for Environmental Prediction Global Forecast System), as well as various regional high-resolution models, have become increasingly mature in capturing large-scale circulation patterns (Lavers, Harrigan et al. 2021), their direct outputs of Quantitative Precipitation Forecasts (QPF) still exhibit significant biases in predicting the intensity and spatial distribution of extreme precipitation. These biases are typically manifested as a generally conservative estimation of heavy precipitation intensity (i.e., systematic underestimation)(Liu, Sun et al. 2021), as well as spatial displacement of precipitation bands (i.e., phase errors) (Kato, Shimizu et al. 2024).

In the field of precipitation forecast post-processing, statistical correction methods such as Bayesian Model Averaging (BMA) (Fraley, Raftery et al. 2010) and Frequency Matching (FM) (Zhu and Luo 2015) have been widely applied. However, these traditional approaches are mostly based on pointwise or local statistical characteristics for value correction and often neglect the spatial structure of precipitation fields (Li, Pan et al. 2022, Liu, Lou et al. 2023). For example, when the precipitation rainband predicted by numerical models is spatially displaced by 1-2 degrees of latitude relative to observations, conventional pixel-level correction methods not only fail to effectively correct the positional bias but may also become ineffective due to the double-penalty effect(Zhao and Zhang 2018, Chen, Sun et al. 2022).

To overcome the limitations of a single model, multi-model blending algorithms have become a key technique for improving forecasting performance (Swinbank, Kyouda et al. 2015, Hamill, Engle et al. 2017). Traditional multi-model blending methods, such as weighted averaging and optimal selection-based blending(Cao, Zhang et al. 2022), can to some extent correct biases in the predicted precipitation location. This correction is essentially achieved through the linear superposition of precipitation fields from different forecasts (Wang, Dai et al. 2021).

Compared with point-wise weighted blending, deep learning techniques, represented by Convolutional Neural Networks (CNNs) (Badrinath, Delle Monache et al. 2023), can more effectively capture the spatial structure of precipitation fields(Li, Li et al. 2024). To this end, we attempt to leverage the global receptive field of U-Net (Ronneberger, Fischer et al. 2015) to construct a multi-model blending model.

However, existing deep learning–based precipitation blending studies still face two key challenges.

First, the distribution of precipitation data exhibits a highly skewed characteristic (Scheuerer and Hamill 2015), with a large number of zero-value samples and very few extreme high-value samples (Dzupire, Ngare et al. 2018, Maity, Suman et al. 2019). Due to this extreme imbalance in the sample distribution (Yu, Sun et al. 2026), directly applying end-to-end regression training often leads to models that, although performing well in terms of mean squared error, fail to preserve the peak intensity of heavy precipitation centers, resulting in a blurring effect (Han, Im et al. 2023).

Second, most existing deep learning models rely solely on gridded reanalysis datasets, such as ERA5((C3S) 2017), as training ground truth. Although gridded data provide good spatiotemporal continuity, the incorporation of smoothing algorithms during the reanalysis data assimilation process often results in the representation of extreme precipitation intensity being lower than that observed at station measurements (Donat, Sillmann et al. 2014, Gervais, Tremblay et al. 2014). This underestimation in observational ground truth (Pang, Shi et al. 2021, Pang, Gu et al. 2025) limits the model's ability to capture extreme precipitation, causing it to systematically underestimate precipitation intensity in operational applications.

To address the above bottlenecks, this study designs a two-stage multi-model blending framework based on an U-Net architecture. The framework transforms the precipitation forecasting problem from a single numerical regression into a two-stage process of "probabilistic classification first, intensity matching second" (Tao, Hsu et al. 2018).

The intensity probability prediction network (named as U-Net G&S) identifies the spatial distribution of the intensity of precipitation, thereby avoiding the extreme smoothing and image blurring caused by direct regression. In addition, a station-grid joint supervision mechanism is adopted. Specifically, real-time observations from 2411 national meteorological stations in China are incorporated as constraints into the loss function of U-Net, compensating for the deficiency of gridded analysis in representing extreme magnitudes.

## 2. Data and Methods

### 2.1. Data

The data used in this study include multi-model gridded forecast data, gridded reanalysis data, and station observation data from 2411 national meteorological stations in China. Among them, the multi-model gridded forecast data serve as the input to the multi-model blending algorithm, while the gridded analysis and station observation data serve as the ground truth for training the neural network.

The gridded data (including multi-model gridded forecasts and gridded reanalysis data) are uniformly interpolated to a grid field covering 0°N-60°N and 70°E-140°E with a spatial resolution of 0.1° × 0.1° (see Fig. 1, which also displays the locations of the 2411 national meteorological stations in China). The interpolation method is the nearest-neighbor approach.

The multi-model gridded forecasts include the 24-hour Quantitative Precipitation Forecasts (QPF) from six models: ECMWF, NCEP-GFS, CMA (China Meteorological Administration)-Meso (Mesoscale Model) (Xu, Wang et al. 2025), CMA-GD (GuangDong) (Zhang, Meng et al. 2024), CMA-SH9 (Shanghai 9KM), and ICON (ICOsahedral Nonhydrostatic Model)(Zängl, Reinert et al. 2015).

The gridded reanalysis data uses the CMPAS-V2.1 (CMA Multi-source merged Precipitation Analysis System) dataset provided by the National Meteorological Information Center (Pan, Gu et al. 2018). CMPAS-V2.1 is an upgraded version of the original China ground-satellite-radar three-source merged hourly precipitation product. On the basis of the original algorithm, a spatiotemporal downscaling algorithm is added to form a three-source precipitation fusion scheme of “PDF (Probability Density Function) + BMA + DS (DownScaling) + OI (Optimal Interpolation)”.

We selected data from 2021 to 2024 to form the training set and validation set. To ensure generalization ability during training and prevent overfitting, 20% of the samples from 2021-2024 were randomly drawn as the validation set for model hyperparameter selection and early stopping decisions; the remaining 80% were used as the training dataset.

Independent observational samples from April to September 2025 (i.e. the flood season in China) served as the test set for final model performance evaluation.

All input variables were standardized.

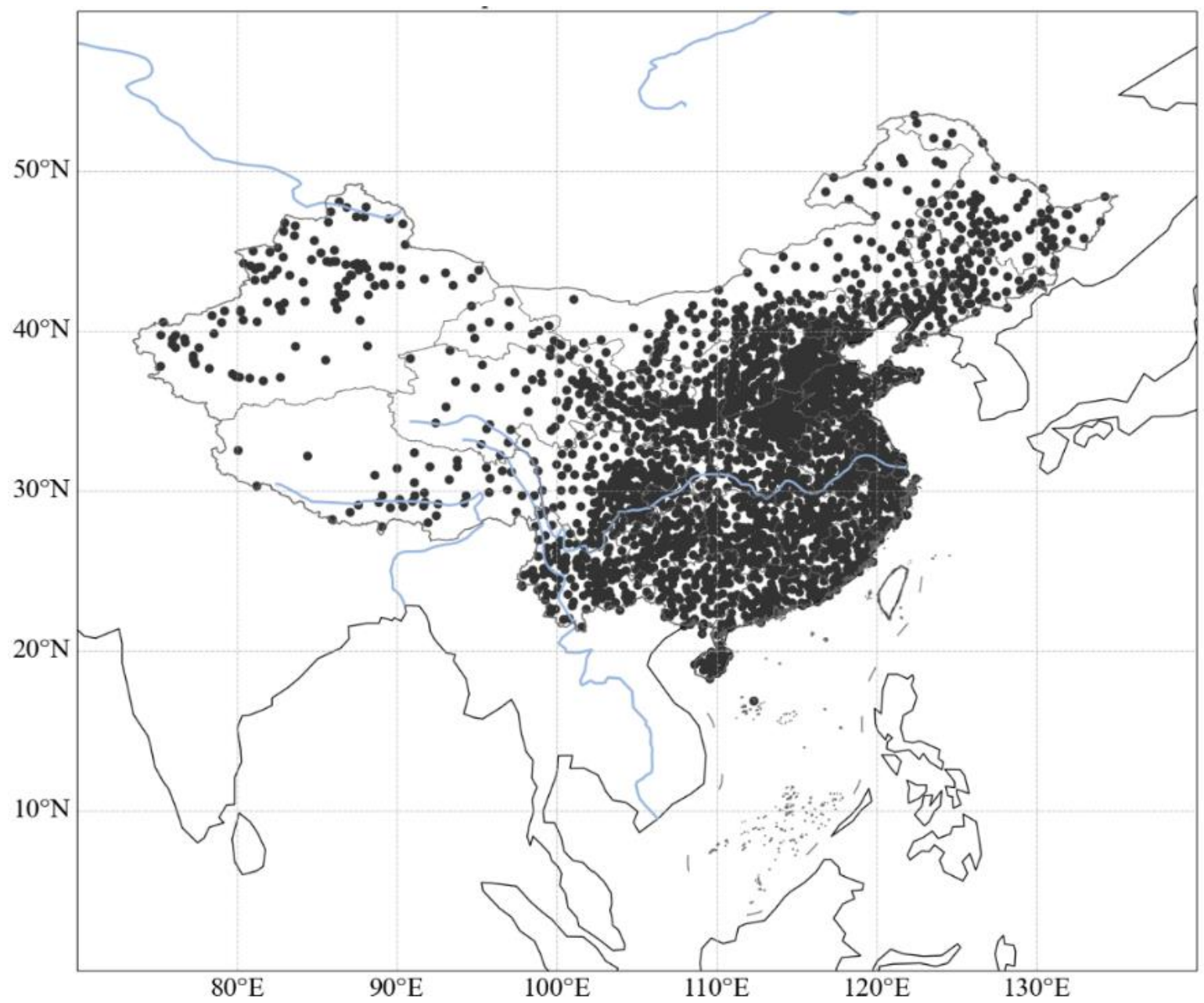

**Fig. 1.** The geographic extent of the gridded analysis data used in this study, along with the spatial distribution of the stations. The black dots in the figure represent the locations of the 2,411 national meteorological stations.

## 2.2. Method

Our study designs a two-stage deep learning-based framework (see Fig. 2), which decomposes the blending process into two core components: (1) a precipitation intensity probability prediction network based on the U-Net architecture; (2) a quantitative field generation algorithm based on probability-intensity conversion and probability matching algorithms.

The adoption of this two-stage structure, rather than an end-to-end regression of QPF directly using a U-Net network, is based on the following considerations:

(1) the distribution of precipitation samples follows a shifted-gamma distribution (Scheuerer and Hamill 2015, Martinez-Villalobos and Neelin 2019), and the sample sizes vary greatly across different precipitation magnitudes. Direct regression learning can easily lead to a severe underestimation of heavy precipitation;

(2) direct regression tends to produce blurry features in the generated precipitation fields(Ravuri, Lenc et al. 2021), resulting in a loss of the textural structure of the precipitation field;

(3) in current operational practice, the evaluation of precipitation forecast accuracy is still primarily based on the Threat Score (TS), and the two-stage scheme can effectively ensure optimization of the TS.

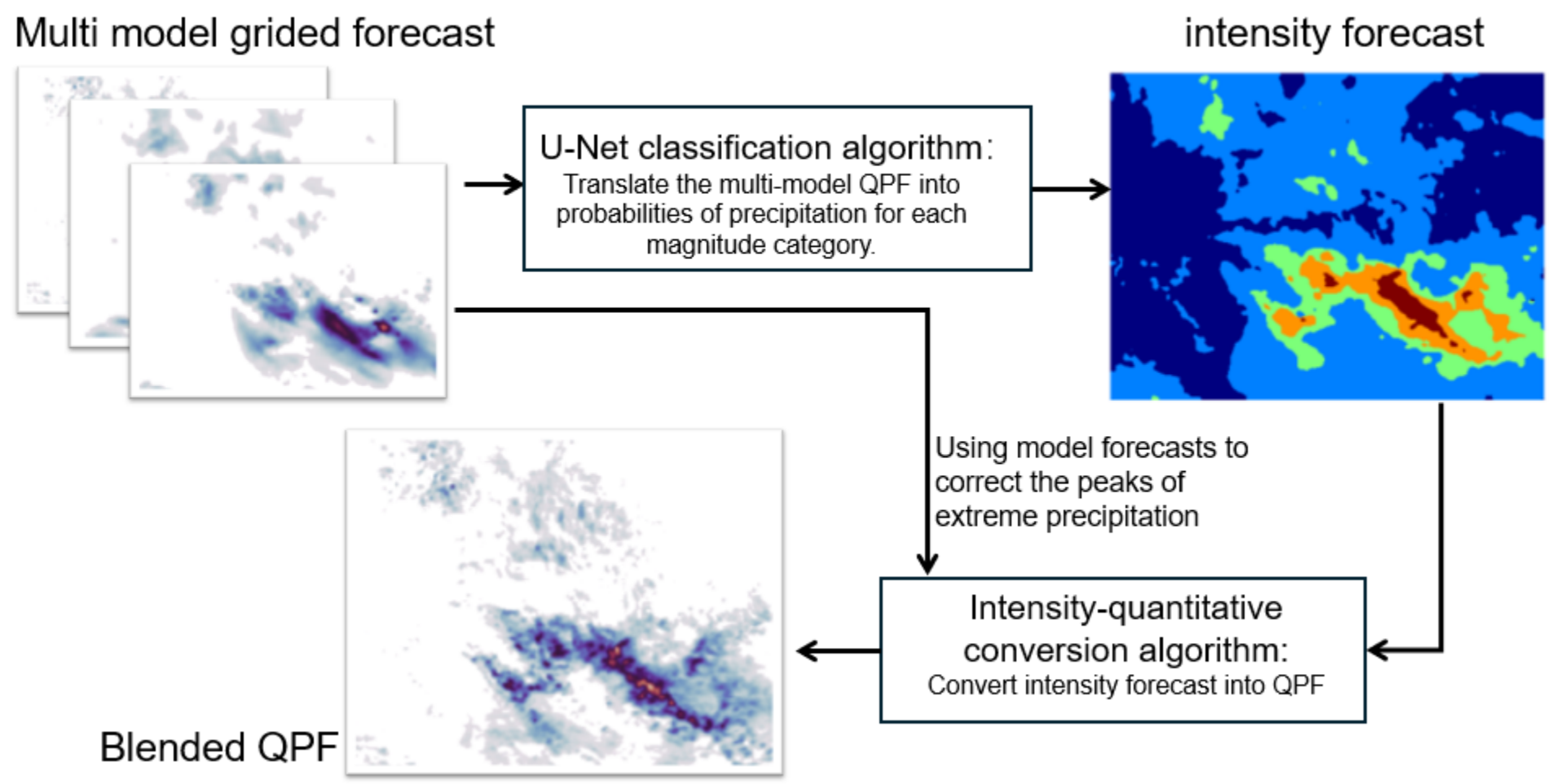

**Fig. 2.** Network architecture of the U-Net based multi-model blending algorithm. The input to the algorithm is the gridded precipitation forecasts from six models. The output of the U-Net classification algorithm is the probabilities of six precipitation intensity categories. The output of the intensity-quantitative conversion algorithm is the blended QPF field.

The precipitation intensity thresholds can be found at table 1. In China's operational meteorology, 24-h precipitation ≥ 50 mm is defined as rainstorm, while ≥ 100 mm is defined as torrential rain. The 50 mm threshold typically corresponds to approximately the 95th–99th percentile of daily precipitation in eastern China, but exceeds the 99.9th percentile in arid regions of western China(Zhai, Zhang et al. 2005).

Therefore, this study adopts a two-stage strategy, transforming the highly nonlinear numerical regression problem into a more stable probabilistic prediction problem. This approach not only alleviates the extreme smoothing issue caused by data imbalance but also enhances the ability to capture the centers of heavy precipitation.

**Table 1.** Categories and threshold ranges for different precipitation intensity levels, along with the corresponding representative precipitation value $R_i$. The representative precipitation value is used in the intensity-quantitative conversion algorithm.

| intensity | No precipitation | Light rain | Moderate rain | Heavy rain | Rainstorm | torrential rain |
| --- | --- | --- | --- | --- | --- | --- |
| Threshold range | <0.1 | [0.1, 10.0) | [10.0, 25.0) | [25.0, 50.0) | [50.0, 100.0) | [100.0, ∞) |
| $R_i$ (mm) | 0.05 | 5.05 | 17.5 | 37.5 | 75 | 200 |

### 2.3. Classification algorithm: U-Net G&S (Grid-station jointly supervised U-Net)

A detailed description of the U-Net network architecture, along with the training parameters, is provided in Supplementary Material 1. Here, we discuss only the "station-grid" joint supervision mechanism in detail.

In operational practice, the evaluation of precipitation forecast accuracy is based on station observations. Meanwhile, gridded reanalysis products suffer from a certain degree of underestimation of actual precipitation. Relying solely on gridded reanalysis as the supervisory ground truth for training leads to an underestimation of heavy precipitation.

Therefore, incorporating station observations to achieve joint station-grid supervision can effectively leverage the high-density spatial information from grids and the high-precision observational advantages of stations, thereby reducing forecast biases in precipitation intensity and location.

In terms of specific implementation, we incorporate the softDice loss (Milletari, Navab et al. 2016) with joint station-grid constraints into the loss function to achieve station-grid joint supervision:

$$\begin{aligned} \mathcal{L} &= \sum_{i=1}^{6} w_i (\lambda_g \mathcal{L}_{Dice}^{g,i} + \lambda_s \mathcal{L}_{Dice}^{s,i}) + \lambda_{L1} \|\theta\| \\ \mathcal{L}_{Dice}^{g,i} &= 1 - \frac{2 \times \sum_m (p_m \times g_m)}{\sum_m p_m + \sum_m g_m + \varepsilon} \\ \mathcal{L}_{Dice}^{s,i} &= 1 - \frac{2 \times \sum_n (p_n \times g_n)}{\sum_n p_n + \sum_n g_n + \varepsilon} \end{aligned} \quad . \qquad (1)$$

In the above equation, $L$ is the loss function of the entire U-Net G&S network, and $\mathcal{L}_{Dice}^{g,i}$ and $\mathcal{L}_{Dice}^{s,i}$ are the softDice losses computed based on gridded reanalysis and station observations, respectively. The superscripts $g$ and $s$ denote the loss terms calculated from gridded reanalysis and station observations, respectively; the superscript $i$ denotes the precipitation intensity category (see Table 1); the subscripts $m$ and $n$ represent each grid point in the grid field and each of the 2411 national meteorological stations, respectively.

To alleviate the class imbalance problem, the weight $w_i$ for each precipitation intensity category is set inversely proportional to the occurrence frequency of that category. In addition, an L1 norm regularization term $\|\theta\|$ is added to the loss function to suppress model overfitting and improve generalization ability.

$P$ (Prediction) is the probability of precipitation occurrence output by the network; its pixel value represents the probability of precipitation at that point (the closer the value is to 1, the higher the probability of precipitation). $G$ (Ground Truth) is the binary map of actual observations, where 1 indicates precipitation occurrence and 0 indicates no precipitation. $\varepsilon$ is a smoothing term, typically set to $1\times10^{-5}$, to prevent the denominator from being zero.

Geometrically, the term $\Sigma(p\times g)$ in Equation (1) actually represents the intersection area of the forecast rainy region and the observed rainy region; the terms $\Sigma(p)$ and $\Sigma(g)$, represent the total expected precipitation area from the forecast and the total observed precipitation area, respectively.

2.4. Intensity-quantitative conversion algorithm

The following equation (2) is used to convert the probabilities of each intensity category predicted by the U-Net G&S into a quantitative value:

$$R = \sum_{i=0}^{6} p_i R_i \; . \tag{2}$$

In the above equation, $p_i$ indicates the normalized probability of precipitation category $i$ predicted by the intensity prediction network (i.e., the output of the U-Net network), and $R_i$ is the representative precipitation value corresponding to that category. The representative value for different precipitation categories are shown in Table 1.

Since the basic intensity-quantitative conversion algorithm using equation (2) above has an upper limit capped at 200 mm (which corresponding to the representative value of torrential rain), which cannot reflect actual extreme precipitation conditions.

Thus, for grid points predicted by the intensity prediction network as belonging to the torrential rain, a probability-matching averaging method (Ebert 2001, Dai, Zhu et al. 2018) is further applied to replace the extreme precipitation value computed by equation (2) with the extreme precipitation value from the ECMWF forecast.

For grid point $m$ predicted by the intensity prediction network as belonging to the torrential rain category, the following formula is used for intensity-quantitative conversion:

$$\begin{aligned} idx_m &= F_{blend}^{-1}(R_m) \\ \overline{R_m} &= \max[100, F_{ECMWF}(idx_m)] \end{aligned} . \quad (3)$$

In the above formula, $R_m$ is the estimated precipitation value at point $m$ calculated by equation (2); $F_{blend}$ represents the sequence obtained by sorting all precipitation values $R$, which is calculated by equation (2), from high to low; $F^{-1}{}_{blend}(R_m)$ represents obtaining the index position of the value $R_m$ in the entire $F_{blend}$ sequence; $F_{ECMWF}$ represents the sequence obtained by sorting all precipitation values from the original ECMWF forecast from high to low; $F_{EMCWF}(idx_m)$ means obtaining the value at location $idx_m$ in the $F_{ECMWF}$ sequence.

### 2.5. Evaluation method

To quantitatively evaluate the performance of our multi-model blending algorithm, and to focus on the contribution of station observation data, we designed the following experimental scheme:

(a) U-Net G (grid supervision group): Only the CMPAS-V2.1 gridded reanalysis is used as the supervision label, i.e., in the loss function (Equation [1]), only the grid-based loss term is retained;

(b) U-Net G&S (joint supervision group): On the basis of the gridded analysis constraint, station observation data from 2411 national meteorological stations are introduced as joint supervision labels, i.e., both the grid-based and station-based loss terms are used;

(c) Baseline comparison group: The six original numerical models participating in the blending, along with the current operational Multi-model Adaptive Integration Technology (MAIT) of the National Meteorlogical Center (NMC) of CMA (Cao, Zhang et al. 2022), are selected as baselines to evaluate the improvement of the our algorithm over point-wise blending methods and single models.

In terms of evaluation metrics, we use the Threat Score (TS) and Bias Score (BS). Additionally, in the case study in Section 3, the Probability of Detection (POD) metric is also used (Cao, Chen et al. 2025). The observational data used for verification are the precipitation observations from 2411 national meteorological stations in China.

The calculation formulas for TS, BS, and POD are given in Equation (4). For different precipitation intensity levels, the corresponding thresholds for True Positives (TP), False Positives (FP), and False Negatives (FN) are calculated separately. The thresholds for each intensity level can be found in Table 1.

$$
\begin{aligned}
TS &= \frac{TP}{TP+FP+FN} \\
BIAS &= \frac{TP+FP}{TP+FN} \\
POD &= \frac{TP}{TP+FN}
\end{aligned}
\quad . \qquad (4)
$$

Taking the rainstorm category as an example:

(1) TP: When the model output precipitation is ≥50 mm, and the actual observed precipitation is also ≥50mm;

(2) FP: When the model output precipitation is ≥50 mm, but the actual observed precipitation is < 50 mm;

(3) FN: When the actual observed precipitation is ≥ 50 mm, but the model output precipitation is < 50 mm.

## 3. Results

### 3.1. Performance improvement via joint supervision

Table 2 presents the TS and BS for 24-hour QPF from multiple models during April-September 2025. ECMWF, NCEP-GFS, ICON, CMA-Meso, CMA-SH9, and CMA-GD are the original NWPs; MAIT is the current operational product of the NMC of CMA; U-Net G and U-Net G&S are the forecast results from U-Net models trained using only gridded analysis and jointly using gridded plus station observations, respectively.

From table 2, it can be seen that among the various products, the U-Net G&S model trained with both gridded and station observations achieves the best performance. Compared with the numerical models, the U-Net G&S model shows significant improvements across all precipitation intensity levels. For the categories ranging from light rain to torrential rain, the TS improvements of the forecast product generated by the U-Net G&S model relative to the ECMWF model are 27.0%, 10.9%, 20.1%, 61.2%, and 140.5%, respectively. Among these, the improvements for the rainstorm and torrential rain are the most notable.

**Table 2.** TS and BS for 24-hour QPF from April to September 2025, verified using station observations.

| | Threat score | | | | | Bias score | | | | |
|---|---|---|---|---|---|---|---|---|---|---|
| models | 0.1mm | 10mm | 25mm | 50mm | 100mm | 0.1mm | 10mm | 25mm | 50mm | 100mm |
| ECMWF | 0.522 | 0.423 | 0.303 | 0.152 | 0.042 | 1.707 | 1.228 | 0.818 | 0.413 | 0.123 |
| NCEP-GFS | 0.543 | 0.402 | 0.291 | 0.170 | 0.057 | 1.738 | 1.248 | 0.726 | 0.513 | 0.270 |
| ICON | 0.577 | 0.425 | 0.307 | 0.169 | 0.054 | 1.603 | 1.135 | 0.793 | 0.503 | 0.201 |
| CMA-Meso | 0.605 | 0.390 | 0.277 | 0.162 | 0.049 | 1.331 | 1.202 | 1.169 | 1.243 | 0.980 |
| CMA-SH9 | 0.591 | 0.375 | 0.256 | 0.148 | 0.049 | 1.471 | 1.414 | 1.652 | 2.097 | 2.164 |
| CMA-GD | 0.614 | 0.392 | 0.285 | 0.177 | 0.059 | 1.268 | 1.202 | 1.318 | 1.484 | 1.158 |
| MAIT[1] | 0.614 | 0.421 | 0.322 | 0.223 | 0.107 | 0.990 | 0.947 | 0.877 | 0.947 | 1.218 |
| U-Net G[2] | 0.625 | 0.457 | 0.349 | 0.222 | 0.094 | 1.286 | 1.147 | 0.846 | 0.732 | 0.611 |
| U-Net G&S[3] | 0.663 | 0.469 | 0.364 | 0.245 | 0.101 | 1.189 | 1.313 | 1.111 | 0.993 | 0.869 |

[1] The current operational multi-model blending product of the NMC of CMA

[2] Using only gridded reanalysis as the training label

[3] Jointly using gridded and station data as the training labels

As another baseline for comparison, compared with the blended precipitation forecast generated by the MAIT, the TS improvements of the precipitation forecast produced by U-Net G&S are 8.0%, 11.4%, 13.4%, 9.9%, and -5.6% for the categories from light rain to torrential rain, respectively. That is, except for the torrential rain category, the U-Net G&S model outperforms the current operational forecast product of the NMC of CMA by a significant improvement across all other intensity levels.

Furthermore, comparing U-Net G and U-Net G&S, we can see that after adding station observations as supervisory data, the TS for each intensity category also show a relatively significant improvement: from light rain to torrential rain, the TS increase by 6.1%, 2.6%, 4.3%, 10.4%, and 7.4%, respectively. From these improvement, it can be seen that adding the station observations has a more pronounced effect on improving the forecast accuracy of heavy precipitation.

### 3.2. Performance for extreme precipitation

Table 3 and 4 present the TS comparisons of various forecast products for the rainstorm (≥50 mm) and torrential rain (≥100 mm) categories, respectively, under the influence of major weather systems in China.

The major weather systems corresponding to each precipitation process are derived from the "Summary Table of Large-Scale Rainfall Events in China 2025" compiled by the duty forecasters at the NMC of CMA.

The participating models include three mainstream numerical models: ECMWF, NCEP-GFS, and CMA-GD, as well as the U-Net G&S algorithm developed in this study.

**Table 3.** TS for rainstorm (≥50 mm) under the influence of different weather systems

| | trough | extratropical cyclone | tropical cyclone | Subtropical high | low-level shear | low-level jet |
|---|---|---|---|---|---|---|
| ECMWF | 0.120 | 0.200 | 0.236 | 0.154 | 0.173 | 0.099 |
| NCEP-GFS | 0.142 | 0.193 | 0.253 | 0.149 | 0.179 | 0.113 |
| CMA-GD | 0.151 | 0.212 | 0.227 | 0.170 | 0.192 | 0.161 |
| U-Net G&S | 0.214 | 0.284 | 0.323 | 0.233 | 0.257 | 0.195 |

As shown in Table 3, for the rainstorm level, the forecast performance of the U-Net G&S model is significantly better than the raw outputs of all evaluated numerical models. The forecasts of the numerical models for rainstorm have generally reached an operationally usable level and exhibit relatively balanced performance under most weather systems.

However, under low-level jet (LLJ) systems, a clear performance divergence is observed among the models: specifically, the TS of the CMA-GD model under low-level jet conditions is markedly superior to those of ECMWF and NCEP-GFS. This may be because the CMA-GD model was developed by the Guangzhou Institute of Atmospheric Science of the CMA, which is targeting the regional characteristics of South China. It possesses stronger dynamic representation capabilities and more refined parameter adaptation for low-level jet systems, which frequently occur in southern China, thereby demonstrating unique advantages in precipitation forecasting for such systems.

For extreme precipitation forecasts at the torrential rain level (see table 4), the performance of the numerical models varies considerably under different weather systems, with overall forecast skill being relatively low.

**Table 4.** The same as Table 3. But for torrential rain.

| | trough | extratropical cyclone | tropical cyclone | Subtropical high | low-level shear | low-level jet |
|---|---|---|---|---|---|---|
| ECMWF | 0.000 | 0.026 | 0.060 | 0.026 | 0.021 | 0.000 |
| NCEP-GFS | 0.016 | 0.046 | 0.157 | 0.048 | 0.038 | 0.009 |
| CMA-GD | 0.017 | 0.071 | 0.127 | 0.061 | 0.054 | 0.011 |
| U-Net G&S | 0.040 | 0.113 | 0.199 | 0.112 | 0.089 | 0.016 |

In typhoon-related precipitation forecasts, although NCEP-GFS and CMA-GD show some forecast potential (TS score > 0.1), the U-Net G&S algorithm proposed in this study achieves a further breakthrough, raising the TS to 0.199 (a 26.8% improvement over NCEP-GFS), demonstrating higher forecast reliability.

More importantly, for extreme heavy precipitation induced by extratropical cyclones and subtropical high, the raw TS of models are all below 0.1, making it difficult to meet the reference needs of operational forecasting. In contrast, the U-Net G&S algorithm successfully raises the TS score for these two weather systems above 0.1, achieving a leap from “low reference value” to “having certain practical utility”.

Although for more complex systems such as troughs, low-level shear, and low-level jets, the TS of both the models and our algorithm remain below 0.1, U-Net G&S still shows substantial gains over the raw models, demonstrating its strong correction capability.

### 3.3. Capability for torrential rain induced by tropical cyclone

Tropcial cyclone (TC) precipitation is a complex weather process driven by the interaction of multi-scale circulations and underlying surface forcing. Due to its extremely complicated physical mechanisms and the high difficulty of forecasting, refined forecasting research on typhoon precipitation has always been a core challenge in both operational meteorology and scientific research (Chen, Guo et al. 2016). Therefore, we select a typical TC case to analyze the correction effect and forecast improvement of our algorithm on TC precipitation.

Considering the frequency of torrential rain (≥100 mm) is relatively high in the TC environment, this study not only evaluates the TS, BS, and POD for the rainstorm (≥50 mm), but also provides the evaluation metrics for the torrential rain.

TC "Co-may" (No. 8 in 2025) featured multiple landfalls, a complex track, and severe wind and rain impacts. Fig. 3 shows the comparison of precipitation forecasts from various models with observations during its landfall in Shanghai on July 30, 2025.

The NCEP-GFS model (Fig. 4a) shows a good forecast baseline for this typhoon, with TS for rainstorm and torrential rain reaching 0.878 and 0.451, respectively. However, its BS for torrential rain is low (0.523), and the POD is also unsatisfactory. This reflects that the numerical model exhibits significant underestimation of magnitude and systematic missed events for extreme heavy precipitation.

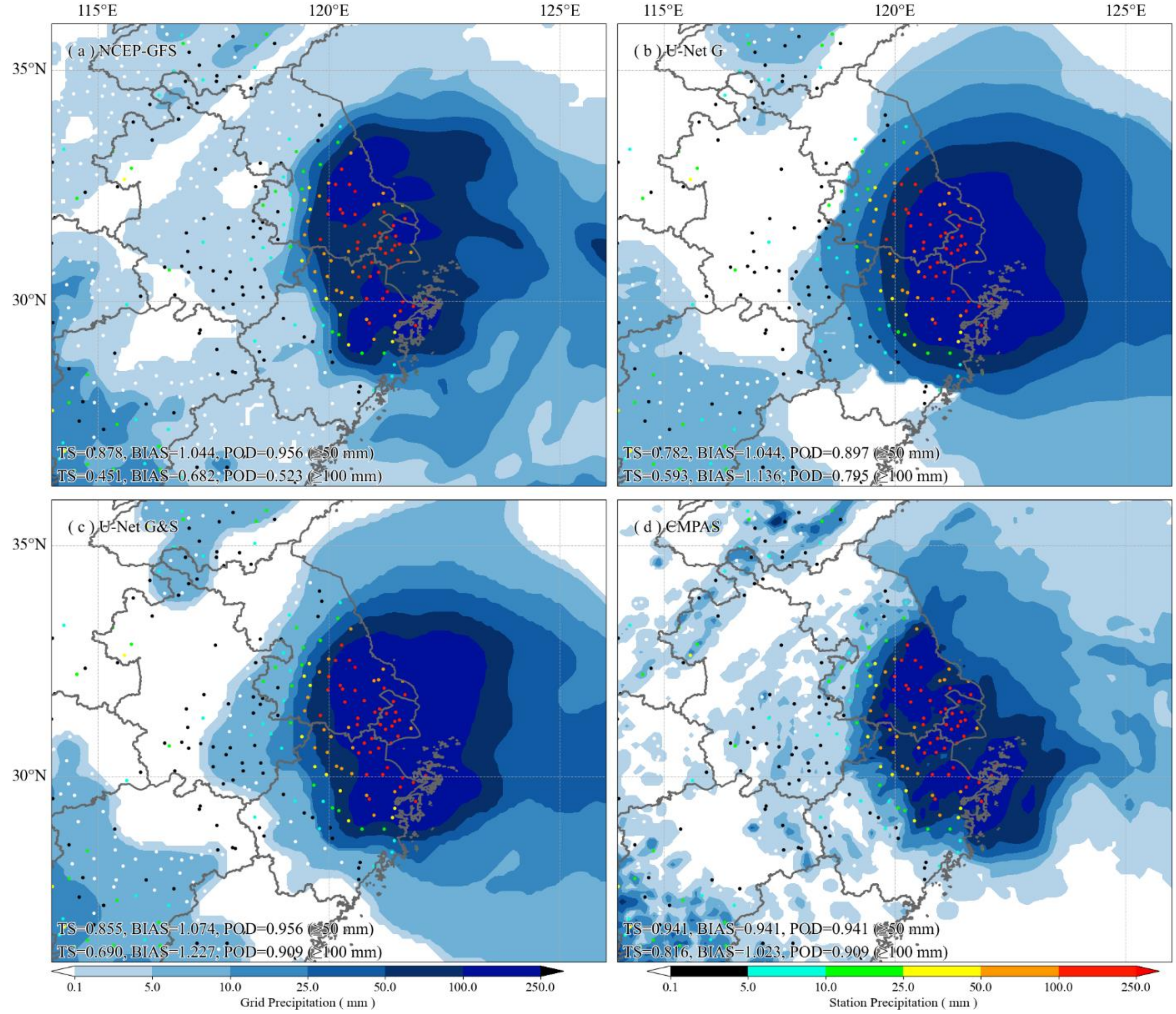

**Fig. 3.** A 24-hour forecast case initialized at 12:00 UTC on July 29, 2025. (a) is the precipitation field forecast by NCEP-GFS; (b) is the forecast field generated by the U-Net G model; (c) is the forecast field generated by the U-Net G&S; (d) is the gridded reanalysis precipitation field. The colored areas represent the gridded forecast precipitation; the dots indicate the actual precipitation observed at stations. The text at the lower left of each panel shows the TS, BS, and POD calculated based on station observations.

In contrast, our blending forecast (Fig. 4c) shows remarkable performance improvement: its TS for torrential rain increases to 0.690, and the BS is close to 1.0. Most critically, this algorithm raises the POD for torrential rain to 0.909, achieving extremely high forecast capability for extreme heavy precipitation events in this case.

### 3.4. Ability to adjust the spatial position of precipitation bands

The two cases compared in this section are a regional precipitation event that occurred on June 26, 2025, in the South China (see Fig. 4), and a precipitation event that occurred on July 21, 2025, in the North China (see Fig. 5).

The main structure of the precipitation band in both events was a narrow, elongated rainband oriented in a northeast-southwest direction. For this type of precipitation, the model's raw forecast products are prone to positional forecast deviations. Such deviations are primarily caused by inaccuracies in the model's forecast of the weather system's location.

As shown in Fig. 4a, for the precipitation event on June 26, the ECMWF model forecast exhibits a pronounced systematic bias. The rainband of rainstorm (≥50 mm) forecasted by ECMWF is displaced northward by approximately 1-2 degrees of latitude overall, and the intensity is generally underestimated. Consequently, the TS for rainstorm (≥50 mm) is merely 0.048, indicating severe underforecasting (misses). Furthermore, the ECMWF forecast produces a relatively large area of spurious precipitation signals from Jiangnan to South China, further degrading the usability of the forecast.

The blended product (Figs. 4b and 4c) exhibits significant improvements. The U-Net model reasonably adjusts the rainband position southward, aligning it closely with the observed heavy precipitation area, while preserving the continuity and structural integrity of the overall rainband pattern.

The model also enhances precipitation intensity, bringing the forecast coverage of rainfall at rainstorm (≥50 mm) and above into closer agreement with observations. Through these corrections, the TS and POD for ≥50 mm rainfall in the U-Net G&S model increase substantially to 0.481 and 0.718, respectively.

Furthermore, regarding the ECMWF false alarm issue over southern China, the model eliminates the spurious signals of widespread light-to-moderate precipitation through multi-model synergistic constraints, thereby improving the spatial accuracy of the forecast.

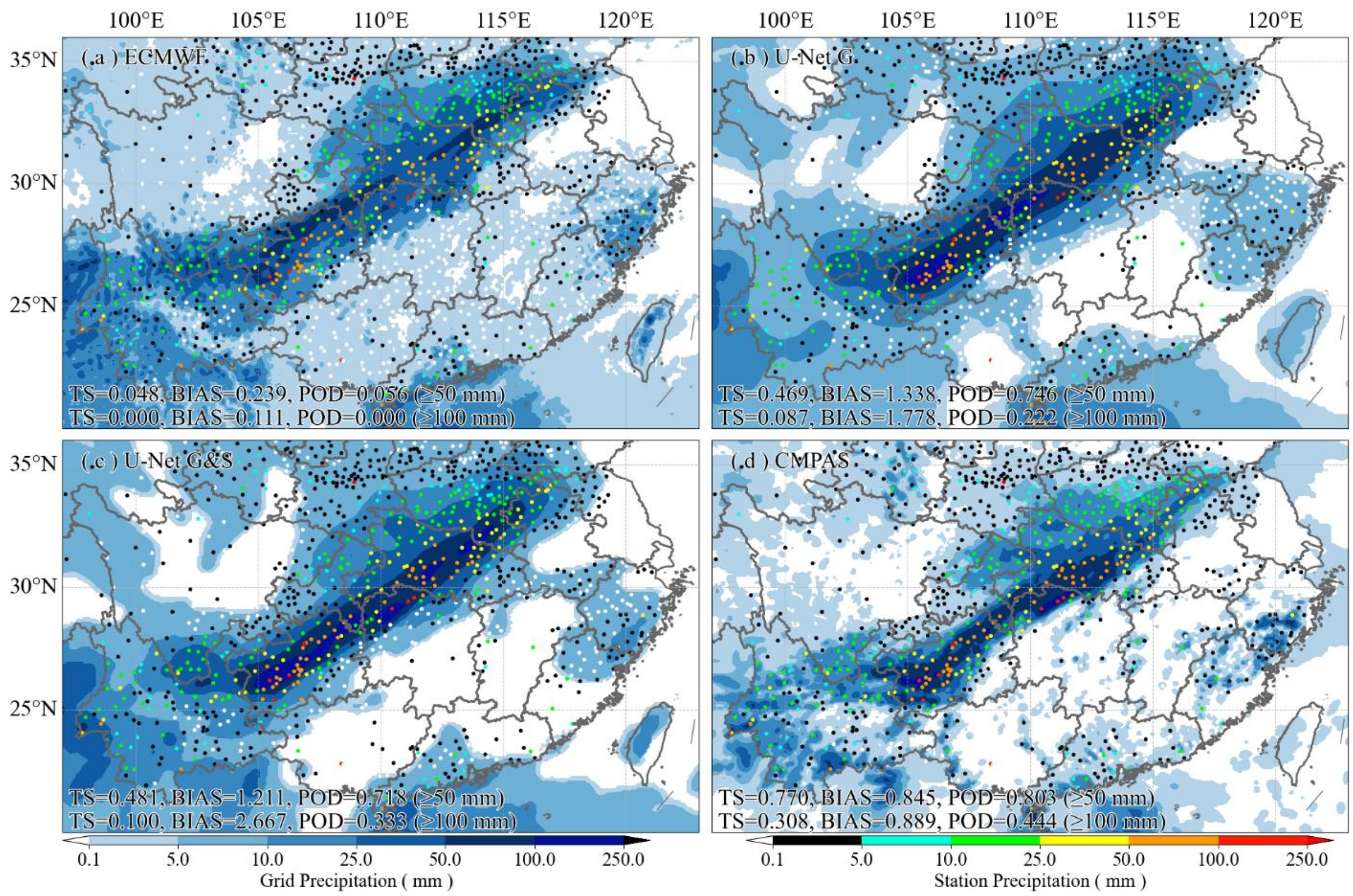

**Fig. 4.** Same as Fig. 3, but for the forecast case initialized at 00 UTC on June 25, 2025, with a 24-hour forecast lead time, and (a) is ECMWF forecast.

For the precipitation event on July 20 (see Fig. 5), the ECMWF model forecasted the spatial position of the rainband with relatively high accuracy (Fig. 5a), indicating that the ECMWF model captured the large-scale circulation pattern of this event reasonably well (the TS for ≥50 mm was 0.250). However, the forecast intensity for rainfall at the ≥50 mm level and above remained conservative (the BS was only 0.486).

Our blended products (Figs. 5b and 5c) rectify the aforementioned intensity bias. While fully inheriting the accurate large-scale rainband position forecast from ECMWF, the algorithm systematically and reasonably enhances precipitation intensity by integrating multi-model intensity information. As a result, the TS for rainstorm (≥50 mm) from the U-Net G and U-Net G&S models improve to 0.309 and 0.443, respectively, representing a marked improvement over the ECMWF forecast.

On the other hand, the BS for the U-Net G and U-Net G&S models are 0.946 and 1.730, respectively. For this case, by substantially increasing the false alarm ratio (as reflected by a BS notably greater than 1), the U-Net G&S model improved the TS and POD by 43.4% and 82.6%, respectively, compared to the U-Net G model.

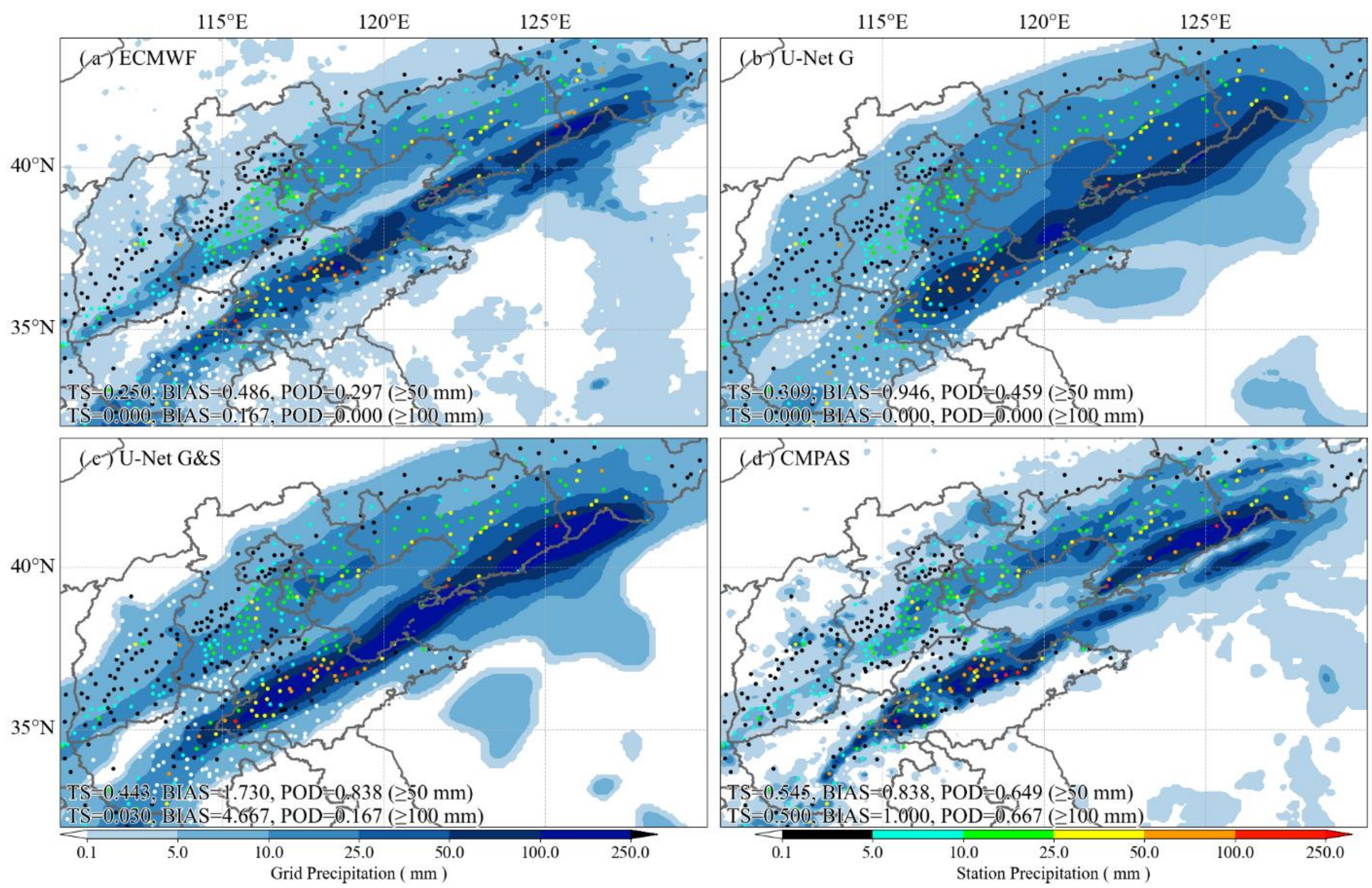

**Fig. 5.** Same as Fig. 3, but for the forecast case initialized at 00 UTC on July 20, 2025, with a 24-hour forecast lead time, and (a) is ECMWF forecast.

## 4. Discussion

### 4.1. Role of station observation data in training

In Table 2, the relatively low BS for U-Net G can be attributed to the fact that, while the classical Dice loss (You, Liang et al. 2023) effectively mitigates the class imbalance issue between positive and negative samples, its penalty mechanism for false alarms and misses is asymmetric.

Empirical evidence indicates that models trained with Dice loss tend to adopt a conservative forecasting strategy to maximize the overlap ratio (Abraham and Khan 2018, Bertels, Robben et al. 2021), resulting in BS significantly below unity for heavy precipitation events.

U-Net G&S was also trained using Dice loss, yet its BS remained close to 1. We speculate that this may be attributable to the following: The gridded observations are derived from the CMAPS 2.1 dataset and are generated through the merging of ground-based, satellite, and radar data, considerations such as algorithmic stability may result in spatial discrepancies between the location of heavy precipitation in the gridded reanalysis product and that recorded at station observations.

This phenomenon can be observed in Figs. 3-5d, where the TS and BS are computed using station observations as the ground truth and the gridded reanalysis product as the forecast; both scores are less than 1.

This indicates that the reanalysis product underestimates the true heavy precipitation intensity. Consequently, using both station observations and gridded reanalysis data concurrently as training labels effectively serves to artificially expand the total sample size of heavy precipitation events. This effectively counterbalances, from an alternative perspective, the inherent asymmetry in the penalty mechanism of Dice loss with respect to false alarms and misses, thereby maintaining false alarms and misses at a favorable equilibrium. By forcing the model to learn from un-smoothed station peaks, the joint supervision artificially enriches the long-tail extreme samples, essentially acting as an implicit class-balancing regularizer.

### 4.2. the capability to correct positional biases of the rainband

The two cases presented in Figs. 4 and 5 demonstrate that the adjustments made by the U-Net model to the gridded precipitation field are not merely a single, mechanical bias correction; rather, the model exhibits a certain capacity for selective judgment.

In other words, the model can, to some extent, assess whether the forecasted rainband position from the numerical model is correct, and when a modest positional deviation occurs, it can adjust the rainband location accordingly.

This adjustment capability likely stems from the spatial bias patterns exhibited by the multi-model forecasts under specific conditions within the historical training dataset. These spatial bias distributions are learned by the model and subsequently applied during the actual blending process.

## 5. Conclusion

Our research develops a two-stage multi-model blending framework based on an U-Net architecture. By decomposing the highly nonlinear precipitation regression problem into two sequential stages: (1) intensity probability prediction and (2) intensity-quantitative conversion algorithm, the framework effectively mitigates the underprediction of extreme precipitation values.

The core innovation of this study lies in the introduction of a station-grid joint supervision strategy. By integrating high-precision surface station observational constraints directly into the loss function, the model's capacity to capture extreme heavy precipitation is substantially enhanced.

Based on verification against independent samples from the 2025 flood season and case studies of representative synoptic systems, the main conclusions of this study are as follows:

(1) the U-Net G&S algorithm significantly outperforms both the raw numerical model output and the current operational product (MAIT). Compared to the ECMWF, the algorithm yields a TS improvement of up to 61.2% at the rainstorm level. Furthermore, it demonstrates superior performance relative to the MAIT operational product at the rainstorm category and below;

(2) U-Net G&S outperforms the U-Net G model, which relies solely on gridded observations, in terms of both TS and BS. The incorporation of station observations as a joint supervision label effectively expands the effective sample size of heavy precipitation events during the training process, thereby counterbalancing the inherent bias of Dice loss under class imbalance conditions;

(3) for extreme precipitation events (≥100 mm) under cyclonic and subtropical high conditions, the algorithm successfully elevates the TS above 0.1, thereby achieving a qualitative transformation of the forecast product from usable to "have certain reference value";

(4) the framework demonstrates an adaptive capability in modifying the spatial structure of rainbands. It can effectively identify and realign systematic positional biases (e.g., meridional shifts of precipitation centers) inherent in raw NWP forecasts. Furthermore, by leveraging synergistic constraints from multi-model inputs, it effectively suppresses spurious precipitation signals (false alarms).

Although U-Net G&S demonstrates considerable potential for extreme precipitation forecasting, certain limitations exists. Therefore, in future work, we plan to:

(1) currently, our definition of extreme precipitation relies on the traditional operational thresholds. In the next phase of this work, we plan to refine the classification thresholds of the blending algorithm by incorporating the climatological characteristics of precipitation, with the aim of further enhancing the model's forecasting capability for extreme precipitation events;

(2) using the explainable AI techniques to investigate how the model is able to achieve a certain degree of correction for the spatial biases of rainbands.

## Acknowledgements

This work was supported jointly by Key-Area Research and Development Program of Guangdong Province (20251102), National Key R&D Program of China (2024YFC3013005) and Key Science & Technology Project of Anhui Province (202523p09050001).